# Unification of Balti and trans-border sister dialects in the essence of LLMs and AI Technology

*Muhammad Sharif[1*,2], Jiangyan Yi[1], Muhammad Shoaib[3]*

[1] State Key Laboratory of Multimodal Artificial Intelligence Systems, Institute of Automation, Chinese Academy of Sciences Beijing

[2] University of Chinese Academy of Sciences, Beijing, [3] Department of Urdu, University of Karachi

sharif.muhammad@ia.ac.cn, jiangyan.yi@nlpr.ia.ac.cn, shoaibwali514@gmail.com

## Abstract

The language called "Balti بلتی, Tibetan: སྦལ་ཏི" belongs to the Sino-Tibetan, specifically the Tibeto-Burman language family. It is understood with variations, across populations in India, China, Pakistan, Nepal, Tibet, Burma, and Bhutan, influenced by local cultures and producing various dialects. Considering the diverse cultural, socio-political, religious, and geographical impacts, it is important to step forward unifying the dialects, the basis of common root, lexica, and phonological perspectives, is vital. In the era of globalization and the increasingly frequent developments in AI technology, understanding the diversity and the efforts of dialect unification is important to understanding commonalities and shortening the gaps impacted by unavoidable circumstances. This article analyzes and examines how artificial intelligence (AI) in the essence of Large Language Models (LLMs), can assist in analyzing, documenting, and standardizing the endangered Balti Language, based on the efforts made in different dialects so far.

## 1. Introduction

Humans, like all animals and plants, thrive in diverse ecosystems. By asserting that biodiversity extends beyond species diversity in nature, as well as the diversity of cultures and languages within human societies, researchers aspire to show that biodiversity has both biological and cultural dimensions [1]. There is a possibility that language death does not occur in privileged communities. It occurs in communities that are disadvantaged and disempowered. Stronger nations, and their languages, may not notice the situation because of their social status. Approximately 290,000 Baltis live in the Baltistan division of Gilgit-Baltistan in Pakistan, where they speak Balti as their mother tongue [2]. Very similar to the spoken Balti language, its trans-border sister dialects are spoken and understood in the Tibet region of China, India, Nepal, Myanmar, and Bhutan because of trade routes, connected geography, religio-cultural and historical basis, and migrations. Figure 2 presents the geographic location of these trans-border sister dialects. Evolved from old Tibetan during the 7th and 9th centuries, dialects of Tibetosphere in literature are Sherpa(Nepal), Burmese (Myanmar), Dzongkha (Bhutan), Ladakhi (India), and Balti (Pakistan) [31]. The number of speakers in these regions is in the millions [30].

Protecting these dialects is of chief significance as they are not just ways of communication but repositories of cultural legacy as well [3]. In the face of modernization and globalization, which intimidate the diversity of minority linguistics, and transfer cultural richness to the next generations it is pertinent to preserve and document these dialects given similarities and differences in linguistic and acoustic perspectives. This paper aims to highlight the importance of preserving and unifying the Balti language and its trans-border sister dialects through the application of artificial intelligence AI. The global community today has been advantageous with the essence of large language Models (LLMs), natural language processing (NLP), and automatic speech recognition (ASR) in finding ways to improve and build cultural understanding, unification, and preservation of commonalities in traditions, culture, and linguistics [4]. This article also provides some points to the technologists, leaders, and politicians who make policies to support the unification and revitalization of endangered languages, dialects, and cultures of the Himalayas, particularly the Tibetosphere.

## 2. Literature review

There is a huge literature in classical Tibetan, especially in Buddhist literature, spread across areas of the states around the Tibetan region of China. Literature [29] gives an in-depth knowledge of various aspects of Tibetan languages, yet it lacks sufficient information about Balti as a live dialect of Tibetan mentioning certain circumstances, and linguistic and acoustic patterns. Branches of the Sino-Tibetan family specifically the languages under consideration in this study are presented in Figure. 1.

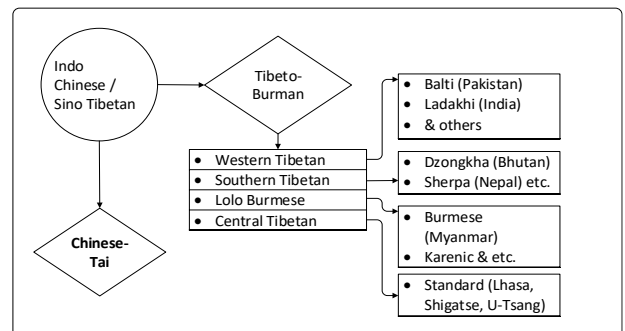

Figure 1: *Position of Balti & Sister Dialects in Sino Tibetan Language Family [28,51]*

---

[*] Corresponding Author

Tibetan is mostly influenced by the Buddhist religious culture of Tibet and one of the main reasons it flourished for centuries to a large region, with classical literature, teaching beliefs by the Monks, and writing scripts found from the past [30, 31]. However, in the last centuries dialects in the neighboring regions, specifically the Baltistan and Kargil regions of the Balti-speaking majority, lost lots of their cultural aspects impacting the pronunciation, writing script, and code-switching with other languages. What the natives of Tibetosphere today behold is the remains from oral traditions and transfer from generation to generation. Particularly focusing on Balti, we observe that Balti communities are fragmented through policies and political boundaries that influence cultural exchange as well as cross-regional communication [13, 14].

Initiatives have been taken by governments, local authorities, and individuals to promote the literature, writing, and teaching of the Balti and the sister dialects in their particular regions. One notable example is the publication of the book "Balti Qaida," in Baltistan. Building on this, the education ministry in Jammu and Kashmir later developed a Balti language curriculum for school students [32]. The Quran and Bible also been translated into Balti. Local poets and writers also been producing literary works in Balti. However, they are all typically written, read, and understood using Perso-Arabic or Roman scripts [33].

Computational studies on Tibetan dialects have extensively covered aspects such as dialect identity, multi-dialect speech recognition, and natural language processing techniques for morphology, translation, and character recognition [34, 35, 36]. Significant attention has also been given to Dzongkha and Burmese [40, 41]. However, this level of computational and technical research is notably absent for dialects spoken outside Tibet, such as those in Nepal, India, and specifically Baltistan. To the best of our knowledge, there has been no computational linguistic work on the Balti language to date. Similarly, dialects like Sherpa and Bodhi in Nepal, Bhutan, and India have primarily been the focus of academic research, particularly concerning their morphological, phonological, and grammatical features, as well as translation and transliteration relationships with other languages [37].

## 3. Linguistic analysis

Balti is a relatively under-researched language in the field of linguistics. However, some existing literature presents root words gathered using corpus data from both documented and naturalistic resources, as well as tense markers [38, 39] and writing styles [33]. Additionally, some books document Balti traditions, grammar, history, and stories [15]. This body of work provides a foundation for further investigation and can be utilized in computational studies of linguistic dynamics. Despite this, there is a notable absence of linguistic and spoken datasets specifically for the Balti language spoken in Baltistan, which poses a challenge for conducting comprehensive research on this language.

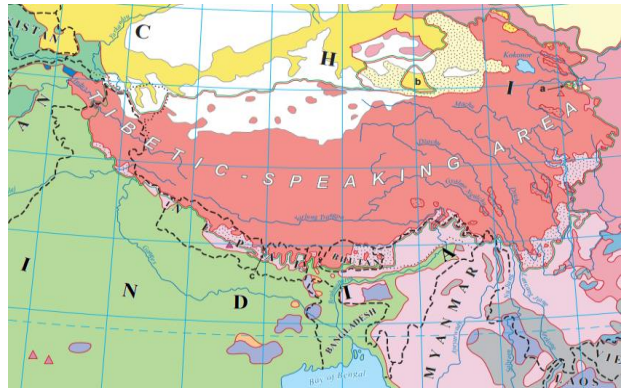

Figure 2: *Geographic Distribution of Tibetan Dialects*

As given in Table 1, it is observed that Mostly nouns are similar in almost all the dialects given. For instance, "meat" is spoken in the similar pronunciation "sha", and when called "Beef" there seem variations, but the native speakers understand what is behind the variation. Like "Bull" or "Ox" is commonly named "Ba" in Balti, Ladakhi, Dzongkha and Sherpa. But "Bull" is also known with another name "Glang, Lang" in these geographies, so the Tibetana and Bhutanese use "Glang+sha" instead of "Ba+sha"; whereas in Burmese, "meat" is pronounced in a similar way "Sha" but the written abugida is "a-sarr", so when its "beef", linguistically "ox or bull meat", it is written with the prefix for "Ox" as "aamell+sarr". It can be concluded that mainly five countries, other than Myanmar, have major similarities in written and spoken forms. Despite different writing ways, pronunciation, and lexical differences, the Tibetan and Burmese have almost similar consonants [20] Table 4 presents a brief sample match of the two sister dialects along with Balti and IPA; likewise, the number system is the same, and even pronounced similarly (from 0 to 9) all across the Tibetosphere, except Bhutan and Burma with few variations from the rest [21]. In Baltistan Balti language is significantly influenced by Urdu and Persian, and interestingly many Balti phonetic dialects have similar effects on these languages locally [26,27], like providing exclusive vowel sounds and softer consonants; however, Tibetan is influenced strongly by Mandarin, and Ladakhi with guttural sounds and tonal variations [17]. So unique phonetic traits are preserved from communities across generations and also reveal socio-cultural distinctions and regional differences. Studies in literature [33, 42, 43], as presented in Table 2, describe variations and similarities in depth.

Balti Language shows important similarities with Sherpa, Ladakhi, and Dzongkha and illuminates common linguistic heritage features more than the dialects in the Tibeto-Burman family. Subject-object-verb(SOV) word orders are followed by these languages uniformly and make sure that verbs are placed at the end of the syntax. However, these languages also use post positions, mean relatable words like "in" and "with" come after nouns [7]. Additionally, the ergative-absolutive case marking system is used for the Sherpa, Ladakhi, Dzongkha, and Balti, where the object of a transitive verb and the subject of an intransitive verb are treated differently from a transitive verb. By the use of prefixes or suffixes, this structural characteristic is more complemented to indicate negation. However, in these languages relative clauses generally come first from the noun they adopt, indicating a pre-nominal configuration [6]. However, in contrast, the researcher Pongsawat (2020) writes in his study that Burmese shows distinct syntactical differences even though it is the part of Tibeto-Burman language family. It is different because it uses a subject-object-verb case marking system rather than an ergative-absolutive system and also positions verbs and SOV word order at the end of sentences. Moreover, pre-nominal relative clauses are followed by Balti languages which are in contrast to the case of clauses in Burmese. As depicted from the samples in Table 3, Burmese shows notably different syntactical structures similar to the other languages due to suffixes they adopt and the pronunciation ways are different as well [5].

Table 1: *Comparing the trans-border dialects [20,5]describe*

| English | Balti بلتی (Pakistan) | Tibetan (China) | Ladakhi (India) | Dzongkha (Bhutan) | Sherpa (Nepal) | Burmese (Myanmar) |
|---|---|---|---|---|---|---|
| **Kings** | Rgyalpoرگیالپو | Rgyalpo རྒྱལ་པོ | rgjalpo | rgyalpo རྒྱལ་པོ | gyalpo | bhurain |
| **Love** | rga رگا | brtseba བརྩེབ | (r)gyal | gaa དགའ་ | gaa | aahkyit |
| **Hunger** | hltoh ہلتوخ | ltogs ལྟོགས | (l)toks | toukê ལྟོགས་ | ltogs | toukê |
| **Meat** | sha شہ | sha ཤ | sha | sha ཤ | sha | a sarr |
| **Beef** | basha با شہ | glangsha གླང་ཤ | basha | langsha,གླང་ཤ | basha | aamellsarr |
| **Fish** | nya نیا | nya ཉ | nya | nya ཉ | nya | ngarr |
| **Mushroom** | shamo شہ مو | shamu ཤེར་མུ | shá-mo | shamu ཤ་མུ | shāmung | mhao |
| **Milk** | oma اوما | oma འོ་མ | oma | Oma འོ་མ | homa | om |

Table 2: *Writing scripts, alphabets, and usage of Balti and Sister Dialects.*

| Writing Style | Characteristics | Alphabet/Script | Usage | Remarks | Writing |
|---|---|---|---|---|---|
| **Perso-Arabic Script** بلتی سکد | This script is taken from Persian فارسی | Alphabets derived from the Urdu | Literature, formal communication | Extensively used in Baltistan, Pak | Written from right to left |
| **Roman Script A-Z.** | This is taken from the English letters A-Z. | It is derived from the Latin Alphabet | For Informal Communication | Transliteration | Written from left to right |
| **Tibetan Script,** བོད་ཡིག | Used for Classical Tibetan. | Classical Tibetan abugida | Literature, and communication | Well preserved in the Tibet, China | Written from left to right |
| **Devanagari** देवनागरी | This script is used for Hindi | Derived from the Hindi Alphabet | Not much common. | Rarely utilized | Written from left to right. |
| **Burmese Abugida** မြန်မာအက္ခရာ | Burmese abugida is used for writing the Myanmar language | From Kadamba or Pallava alphabet of South India | In literature and communication. | The official language of Myanmar | Left to right segmental writing system |
| **Bhutanese (Dzongkha** རྫོང་ཁ**)** | Dzongkha uses the actual Tibetan scripts | Same as Tibetan, cursive long hand | Literature and communication | Bhutan National, official language | Written from left to right |

It is observed that Balti dialects further diversify from speech patterns and regional accents with elongated rhythmic intonation and vowels featured by Baltistan and inclined by Persian poetry, however, Ladakhi exhibits tonal shifts and abrupt consonant clusters similar to traditional Tibetan. The influence of nearby languages like Shina, Kashmiri, and Hindi introduces more phonetic variations, preserving different oral civilizations, and elevating the linguistic landscape and local narratives [18]. Due to diverse linguistic and cultural influences, the Balti dialect's vocabulary varies drastically with words, as the same word in different regions speaks differently [19]. Similarly, the syntax making shows differences as well as similarities due to the position of verbs and nouns. However, by constructing a rich variety of linguistic expressions, familial terms, numerals, and everyday objects used, specifically in Baltistan, with Urdu and English; Balti vocabulary profoundly influences local cultures and languages.

## 4. Unification through technology

### 4.1. Adopting AI Technology

Technological advancements, particularly the advent of artificial intelligence, have made it possible to bridge gaps, unify similarities, and strengthen ties across geo-political, religious, and generational boundaries. In the context of Balti and its sister dialects, there is a pressing need for dialect unification, leveraging the power of technology. Automatic Speech Recognition (ASR) has proven effective in recognizing traditional speech, accents, dialects, and various spoken language varieties. Similarly, Natural Language Processing (NLP) has proven a crucial role for processing and analyzing linguistic perspectives. NLP aids in developing culturally sensitive models, and by utilizing linguistics, statistics, and artificial intelligence, large language models (LLMs) have demonstrated the ability to understand text meanings and mimic human languages [4]. From variations in phonological and phonetics, speech recognition can enhance speech from different dialects.

Table 3: Syntactic *Similarities and Differences*

| Phrases / Dialects | When should I come? | New Year | Yes, please / Thanks |
|---|---|---|---|
| **Balti** | nga nam ong? نَا نام اونگ | lo sar لو سر | ju le , aju le جو لے، اجو لے |
| **Dzongkha** | nga nam ong? ང་ནམ་འོང་? | losa ལོ་གསར་ | kadriche བཀའ་དྲིན་ཆེ། |
| **Burmese** | Be-a-chin-la-ma-la ဘယ်အချိန်လာရမလဲ | hnit tha နှစ်သစ် | chay-zu-tin-ba-te ကျေးဇူးတင်ပါတယ် |
| **Sherpa** | nga nam pheu? | lo sar नयाँ | thuche/horche. धय वाद |
| **Ladakhi** | nga nam yong? | lo sar | ju-le |
| **Tibetan** | Nga nam yong ba? ང་ནམ་ཡོང་བ? | lo gsar ལོ་གསར | thOkce / thuk-je-che ཐུགས་རྗེ་ཆེ། |

Table 4: Sample mapping of Burmese Consonants[12]

| Tibetan | Burmese | Balti(Persian) | IPA |
|---|---|---|---|
| ཀ | က | کا | /ka/ |
| ཁ | ခ | کھا | /kʰa/ |
| ག | ဂ | گا | /g/ |
| ང | ၌ | نا | /n/ |
| ཏ | တ | تا | /t/ |
| ཤ | ဆ | شا | /sʰ/ |
| ཝ | ဝ | وا | /w/ |

Oral culture can be protected and mapped through audio archives to support language learning by having real-time feedback. Tibetan dialects across the formal Tibetan region within China have already worked on speech and linguistic perspectives with great accuracy [34,36]. Likewise, Burmese and Dzongkha (Bhutanese) dialects also got considerable digitization in the past. For instance, corpus development for speech recognition of Dzongkha was found in the literature, word segmentation, and text-to-speech [40,44]. Pre-trained models for the Myanmar language have also been studied in multiple pieces of literature [41], also local dialects of Myanmar have been explored [13]. Similar strategies adopted to match similarities and variations in linguistic and acoustic perspectives across the Balti and other sister dialects can lead to a unified system of dialogue and understanding between the communities spread with geo-political boundaries.

It is observed that large language models (LLMs) are used to teach varied linguistic information, translate text, and facilitate communication across different dialects [22], taking advantage of which, the understanding and communication gap can be addressed by unifying the differences and developing dialect-specific glossaries and dictionaries. A recently published literature [23] demonstrates the dialectic gap in the performance of LLM-based machine translation and automatic speech recognition for several dialects, including Arabic, Mandarin, Finnish, German, Bengali, Tagalog, and Portuguese. Along with the LLMs integrated with ASR and NLP tasks, Multi-Modal techniques have recently made great advancements [39,42], helping in language analysis with the help of text, audio, and visual data to understand dialects. So by using advanced models, technology, and AI we can preserve and unify Balti and sister dialects, enhance cultural heritage, and measure linguistic diversity. It also helps to transfer Balti-speaking culture to future generations.

### 4.2 Strategies for Unification

Different strategies have been applied for unifying dialects, such as creating a standardized dictionary that contains common phrases and words and also has proper meanings and pronunciations of full words present in the dictionary [24]. Similarly, building standard phonetics via technical tools and acoustic analyses [30,25]. However, recognition and writing standards are maintained through unified scripts that have syntactic and phonetic features of all targeted dialects, promoting standard written communication. Transliteration tools and OCR software have been traditional ways of text conversion and recognition and also assist in transmitting the given text into new forms of script and making pronunciation lexicon in speech recognition, identification, and understanding. These approaches altogether support a unified linguistic and cultural uniqueness while protecting the wealthy tradition of the Balti-speaking society.

LLMs assist not only in textual communication mediums, rather it offers great advancements in spoken language understanding, recognition, identification addressing dialectal, age, sex, medium and other variations. Hence the similarities of Balti and sister dialects can also be matched and unified. It is essential to process required speech and text data, for the said tasks, process them using sophisticated algorithms like seq2seq transformer models by the open source multilingual Whisper model (Radford, 2023), and similar level competitive yet effective strategies in Wav2Vec2 by Meta AI (Baevski, 2020).

## 5. Cultural and demographic impacts

Unifying Balti and sister dialects can have a great impact both culturally and demographically. Cultural heritage can be preserved through a unified language, even though traditional information, values, and folklore are constantly acknowledged and transmitted. By unity of the dialects, countries and societies where Balti and sister dialects are spoken and understood can bridge the cultural gap, and have firm bonding and mutual understandings. Enhanced communication within and among these zones simplifies connections in learning about their common history, ecology, environmental effects, common values, and community settings. Furthermore, a united language strengthens communal solidity and uniqueness, promoting a joint cultural inheritance and harmony. This agreement supports local firmness and collaboration, decreasing social division and fostering mutual struggles in addressing challenges, thus contributing to in general regional peace and synchronization.

## 6. Conclusion and prospects

By the fusion of Balti and sister dialects, with the help of technological and AI expansions, major linguistic, historical, and cultural advantages are possibly offered to the speakers of these trans-border sister dialects. This study emphasizes the significance of establishing phonetic norms, consistent dictionaries, and unified scripts to protect cultural heritage and improve communication. Prospect study must center on refining linguistic examination, increasing digital records, and mounting AI-driven apparatus for text generation and dialect normalization. AI has a huge potential in cultural unification and linguistics, giving new solutions to bridge linguistic gaps, promote common understanding, and ensure the protection of varied linguistic civilizations. Implementing and clothing this dream, can enrich our cultural tapestry worldwide and preserve the language of Tibetosphere for future generations.

## 7. Acknowledgements

This work is supported by the National Natural Science Foundation of China (NSFC) (No. 62322120, No. U21B2010, No. 62306316, No. 62206278).